\documentclass[pdflatex,sn-mathphys-num]{sn-jnl}% Math and Physical Sciences Numbered Reference Style 
\usepackage{graphicx}%
\usepackage{multirow}%
\usepackage{amsmath,amssymb,amsfonts}%
\usepackage{amsthm}%
\usepackage{mathrsfs}%
\usepackage[title]{appendix}%
\usepackage{textcomp}%
\usepackage{manyfoot}%
\usepackage{booktabs}%
\usepackage{algorithm}%
\usepackage{algorithmicx}%
\usepackage{algpseudocode}%
\usepackage{listings}%
\usepackage{csquotes}
\usepackage{url}
\usepackage{array}
\usepackage[caption=false]{subfig}
\usepackage{amssymb}
\usepackage{amsmath}
\usepackage{lmodern}
\usepackage{anyfontsize}
\usepackage[table]{xcolor}

\begin{document}

\title[Article Title]{Learning Geometric Invariant Features for Classification of Vector Polygons with Graph Message-passing Neural Network}


\author*[1]{\fnm{Zexian} \sur{Huang}}\email{zexianh@student.unimelb.edu.au}

\author[1]{\fnm{Kourosh} \sur{Khoshelham}}\email{k.khoshelham@unimelb.edu.au}
\equalcont{These authors contributed equally to this work.}

\author[1]{\fnm{Martin} \sur{Tomko}}\email{tomkom@unimelb.edu.au}
\equalcont{These authors contributed equally to this work.}

\affil[1]{\orgdiv{Department of Infrastructure Engineering}, \orgname{The University of Melbourne}, \orgaddress{\street{Grattan Street}, \city{Melbourne}, \postcode{3010}, \state{Victoria}, \country{Australia}}}


\abstract{Geometric shape classification of vector polygons remains a challenging task in spatial analysis. Previous studies have primarily focused on deep learning approaches for rasterized vector polygons, while the study of discrete polygon representations and corresponding learning methods remains underexplored. In this study, we investigate a graph-based representation of vector polygons and propose a simple graph message-passing framework, PolyMP, along with its densely self-connected variant, PolyMP-DSC, to learn more expressive and robust latent representations of polygons. This framework hierarchically captures self-looped graph information and learns geometric-invariant features for polygon shape classification. Through extensive experiments, we demonstrate that combining a permutation-invariant graph message-passing neural network with a densely self-connected mechanism achieves robust performance on benchmark datasets, including synthetic glyphs and real-world building footprints, outperforming several baseline methods. Our findings indicate that PolyMP and PolyMP-DSC effectively capture expressive geometric features that remain invariant under common transformations, such as translation, rotation, scaling, and shearing, while also being robust to trivial vertex removals. Furthermore, we highlight the strong generalization ability of the proposed approach, enabling the transfer of learned geometric features from synthetic glyph polygons to real-world building footprints.}




\maketitle

\section{Introduction}
\label{sec:introduction}
Geometric shape classification of spatial objects is a non-trivial task in spatial analysis. The recognition of spatial objects assisted by automated shape classification is a major enabler of data intensive tasks, including cartographic generalisation, building pattern recognition, archaeological feature analysis, and road geometry identification \citep{yan2019graph,veer2018deep,andravsik2016efficient}. 

One of the main challenges in geometric shape classification is the identification of object footprints (i.e., outlines) in the geographic context. A key requirement is the classification of objects invariant to geometric transformations including rotation, scaling, and shearing. The human visual system relies on the Gestalt properties of perceived objects \citep{Lehar2003-LEHTWI,lehar2003gestalt} during the classification of shapes. A key property of Gestalt is the invariance to geometric transformations, assuring that geometric shapes are recognized regardless of geometric transformations. In contrast to the human visual perception capabilities, deep learning architectures have been designed and shown to be mainly translation-invariant (e.g., Convolution Neural Networks, CNN) or permutation-invariant (e.g., Graph Neural Networks, GNN) on classification tasks. The ability to reflect Gestalt principles through transformation invariances thus present a strong motivation for an inductive bias in the design of learning-based models for geometric shape recognition of spatial objects. 

Spatial objects are often conveniently represented as vector polygons, a discrete data representation thus far neglected in deep learning research. Learning geometric-invariant features from vector polygons has the following requirements: (1) a generic data representation that encodes geometric features of polygons without information loss; and (2) a learning model built on this input data representation that enables learning latent geometric-invariant features robust to geometric transformations. Existing geospatial applications (i.e., shape coding and retrieval \citep{yan2021graph}, building patter recognition \citep{veer2018deep,yan2019graph} and building grouping \citep{yan2022graph}) utilizing polygonal geometries motivate us to systemically study geometry encoding methods in conjunction with appropriate and robust learning architectures that enable learning transformation invariant features of spatial polygon geometries. 

\citet{veer2018deep} proposed VeerCNN, a deep convolution model to learn convolutional features on fixed-size 1D vertex sequences of polygon vertices for building attribute recognition. The architectural limitations of CNNs attributable to shared-weights convolutional kernels followed by non-linear activation and fixed-size pooling layers (e.g, $mean$, $sum$ or $max$) only enable learning intermediate hierarchical features invariant to translation, with poor handling of rotation, scaling and shearing. \citet{mai2023towards} noted that CNN models learning on 1D sequences are also sensitive to (1) permutations of the feed-in order of polygon vertices (i.e., sensitive to \emph{loop origin invariance}); and (2) the impact of trivial vertices on the exterior and interiors of polygons, defined by \citet{mai2023towards} as \emph{``[vertices] \ldots where the addition or removal of the vertex have no effect on the geometric shape and topological properties of the outlines of the polygons''}. The addition or removal of trivial vertices in polygons does not alter the semantic information of the shape (as captured by the human-assigned label), closely aligning with the Gestalt principles.

Learning models tailored for discrete (i.e., not grid-like), permutation invariant  data representations are predicated on data domains where entries hold no explicit neighborhood information (e.g., point sets \citep{zaheer2017deep,qi2017pointnet,qi2017pointnet++,lee2019set}). Polygons can then be encoded into point sets of vertices on the exterior and interior rings of the polygons with arbitrary ordering. In contrast to the 1D sequence encoding of polygons, point set encoding assumes input data to have varying input sizes and feed-in order. Models learning on point sets are designed to map input points with permuted order to task-specific outputs (i.e., \emph{permutation invariance} \citep{bronstein2021geometric}). Here, we argue that such point set representations do not sufficiently capture the connectivity information between vertices of polygons, leading to information loss and performance degradation in geometric shape classification. 

Graph data structures present a suitable data encoding for polygons. Recent studies \citep{yan2021graph,he2018recognition,bei2019spatial} convert vector polygons into un-directed graphs where the vertices on the exteriors and interiors of polygons are captured as graph nodes linked by un-directed graph edges. Compared to fixed-size 1D sequences and varying-size point-set encoding of polygons, graph encoding effectively encapsulates both the geometric and connectivity information of the vertices along the exterior and interior linear rings defining the polygons. Graph-based representations also enable the feed-in order of polygon vertices to be independent of the model outcomes, while the connectivity (topology) between parts of polygon representations remains invariant to geometric transformations.

Graph convolutional autoencoders (GCAE) \citep{yan2021graph} extend graph convolutional neural networks (GCNs) \citep{kipf2017semi} by learning spectral graph embeddings from polygon graphs, demonstrating the effectiveness of graph latent embeddings in polygon shape retrieval. \citet{bronstein2021geometric} noted that spectral graph convolutions aggregate node features from neighboring nodes with constant weights, making them highly dependent on the topological structure of the input graphs. In vector polygons, vertices on linear rings have fixed neighbors (i.e., the left and right adjacent vertices), which reduces the expressivity of graph convolutional features learned from polygon graphs. In this study, we leverage graph message-passing mechanisms \citep{gilmer2017neural} and propose a simple graph message-passing network, PolyMP, along with its densely self-connected variant, PolyMP-DSC, to learn more expressive and robust latent representations of polygons. We hypothesize that combining graph-based representations of polygons with graph message-passing models facilitates learning robust latent features that remain invariant to geometric transformations (i.e., rotation, scaling, and shearing), thereby improving the generalizability of shape classification across datasets.

If our hypothesis holds, we suggest that downstream tasks performed on the learned features of vector polygons will exhibit robust performance on shapes with varying amounts of trivial vertices. We present a series of experiments designed to evaluate model robustness on polygons subjected to geometric transformations and investigate the generalizability of our findings across different combinations of polygon representations and model architectures (permutation-invariant vs. translation-invariant models) for classifying vector polygons with and without holes (inner linear rings).

Following a literature review (Section~\ref{sec:background}), we introduce a synthetic dataset of polygonal shapes with high shape variability and well-defined human labels, derived from Latin alphabet character glyphs (Section~\ref{glyph}). Despite significant shape variations across fonts, particularly in terms of trivial vertices, characters maintain strong recognizability due to their Gestalt properties. We then benchmark the performance of learning models on three distinct data representations (1D sequence, set, and graph) under varying geometric transformations (i.e., rotation, scaling, and shearing). Our results (Section~\ref{results}) document the models' robustness to geometric transformations through performance evaluation on the synthetic glyph dataset. Building on these findings, we further assess model generalizability using a real-world building footprint dataset from OpenStreetMap (OSM) \citep{yan2021graph}.

Our main contributions are: 
\begin{enumerate}
    \item we provide a systematic investigation of vector polygon representations for geometric shape classification, highlighting the strengths of graph-based encoding over sequences and point sets;
    \item the introduction of PolyMP and PolyMP-DSC, two lightweight graph message-passing architectures designed to learn geometric-invariant and robust latent representations;
    \item the release of a synthetic dataset of vector polygons derived from character glyphs, enabling controlled benchmarking of geometric learning models; and
    \item extensive experiments which show that the proposed message-passing models, PolyMP and PolyMP-DSC, significantly improve robustness to transformations and generalize well to real-world geospatial data.
\end{enumerate}

Together, these contributions aim to provide a principled and practical foundation for learning geometric-invariant representations of spatial polygons—pushing forward the capabilities of shape-based analysis in geospatial domains.

\section{Background}
\label{sec:background}

\subsection{Machine Learning with Vector Geometries}
\label{sec:ml_with_vector}

Building on computer vision approaches for image classification, polygonal shape classification has traditionally relied on learning geometric features from rasterized 2D vector geometries using deep CNN architectures \citep{he2016deep,wu2016shape}. The geospatial community has adopted these methods for various remote sensing tasks, including vector shape generation. For example, the Microsoft Open Building Footprints dataset \citep{microsoft} was generated by segmenting and vectorizing building polygons from satellite and aerial imagery using deep semantic segmentation networks \citep{lin2017refinenet}.

This expertise in image-based learning naturally influenced vector shape learning. \citet{xu2017quality} trained a deep convolutional autoencoder to assess the quality of rasterized building footprints collected from OpenStreetMap (OSM) \citep{OpenStreetMap}. Similarly, \citet{veer2018deep} evaluated deep neural networks for classifying spatial vector geometries, encoding geometry vertices as 1D sequences directly fed into deep models. While CNNs and RNNs performed comparably to hand-crafted feature-based methods, they lacked the ability to learn transformation-invariant shape features.

Beyond geospatial applications, typography and computer graphics research also leverage vector shape learning. \citet{lopes2019learned} developed a convolution-based generative model to create vector text glyphs, but it was optimized primarily for scale-invariant font style learning. \citet{mino2018logan} and \citet{sage2018logo} generated vector icons and logos from rasterized images using deep generative autoencoders \citep{DBLP:journals/corr/RadfordMC15, odena2017conditional}. Additionally, \citet{ha2018neural} introduced Sketch-RNN, a recurrent neural network for generating stroke-based vector drawings, while \citet{carlier2020deepsvg} proposed DeepSVG, a hierarchical generative model using Transformer-based encoding for scalable vector graphics (SVG).

\subsection{Vector Geometry as Discrete Data Representation}

Recent advances in graph representation learning \citep{DBLP:journals/corr/BrunaZSL13,defferrard2016convolutional,kipf2017semi,velickovic2017graph} have reformulated vector polygon learning as a graph-based or set-based problem. \citet{yan2019graph} applied graph convolution networks (GCNs) to building pattern classification, structuring buildings into clusters based on geometric relationships using Delaunay triangulation and Minimum Spanning Trees. Their GCN model classified building clusters into regular or irregular patterns. Similarly, \citet{bei2019spatial} introduced a spatially adaptive model using graph encoding for group pattern recognition, where buildings served as nodes in a graph convolutional network (GCN).

\citet{yan2021graph} proposed a graph convolutional autoencoder (GCAE) for building shape analysis, encoding building polygons as graphs where boundary vertices acted as nodes. However, their method was sensitive to  rotations, limiting its robustness for polygon shape retrieval. Addressing this limitation, \citet{huangcontrastive} introduced a contrastive graph autoencoder to enable robust polygon shape matching and retrieval in large-scale vector datasets.

\citet{liu2021triangleconv} developed a deep point convolutional network (DPCN), a modification of a DGCNN \citep{wang2019dynamic}, for building shape recognition in map space. Their model introduced the \texttt{TriangleConv} operator, which extracts local geometric features from triangle-based representations of polygon vertices: $f_{\texttt{TriangleConv}}(x_{i-1}, x_{i}, x_{i+1}) = \{(x_{i} - x_{i-1}), (x_{i} - x_{i+1}), (x_{i-1} - x_{i+1})\}$, where $(x_{i-1}, x_{i}, x_{i+1})$ are adjacent point features of a building polygon. Despite achieving competitive classification performance on building footprints \citep{yan2021graph}, this method requires pre-defining local triangles of polygons and repeatedly computing redundant local geometric features (i.e., angles and areas) of polygons. Furthermore, it does not evaluate the robustness of learned representations to geometric transformations such as rotation, scaling, and vertex perturbations.

\citet{mai2023towards} introduced ResNet1D, a 1D CNN-based polygon encoder that captures local geometric features for shape classification. It encodes polygon vertices using relative offsets from neighboring vertices: $\{x_i, x_{i-1} - x_{i}, x_{i+1} - x_{i}, \cdots ,x_{i+k} - x_{i}, x_{i-k} - x_{i}\}$. They compared this encoding with NUFTSpec, a Non-Uniform Fourier Transform (NUFT)-based method that aggregates global shape features in the spectral domain. While ResNet1D focuses on shape classification, NUFTSpec is more suited for predicting topological relations. Their work aligns with our study in exploring different polygon encoding methods, but our focus is on evaluating transformation-invariant learning architectures across various representations (CNN, set-based, graph-based, and attention-based models).

While previous studies explore graph-based and local geometric encoding for vector polygons, they often lack a principled justification for applying graph learning methods to polygon representations. We address this gap by providing: 1. Experimental validation grounded in theoretical motivation for robust polygon shape learning; 2. A principled framework demonstrating how permutation-invariant learning architectures combined with discrete data representations improve robustness to geometric transformations and vertex perturbations.

\subsection{Graph Representation Learning}
\label{sec:graph_rep_learning}
Graph representation learning model leverages the property of data permutations and learns a whole graph embedding $\mathcal{G}^\star$ from the input graph $\mathcal{G}$ with differentiable permutation-invariant function $\mathcal{F}$. The learned embedding $\mathcal{G}^\star$ can be applied to graph-wise and node-wise classifications, or even edge predictions. 

By representing polygon linear rings as graphs, we can define a differentiable permutation-invariant function $\mathcal{F}$, which takes the graph representations as inputs and returns a corresponding compact graph embedding $\mathcal{F}(\mathcal{G};\theta) := \mathcal{F}(X, A;\theta) \rightarrow \mathcal{G^\star} \in \mathcal{R}$. 

Given a permutation matrix $\mathcal{P}$ acting on a graph $\mathcal{G}$, it reorders the node feature matrix $X$ and the adjacency matrix $A$, producing a permuted graph representation $\mathcal{G}^\prime = (PX, PAP^\intercal) = (X^\prime, A^\prime)$. The ideal permutation-invariant function $\mathcal{F}$ should satisfy $\mathcal{F}(X^\prime, A^\prime;\theta) = \mathcal{F}(X, A;\theta)$. We relate the idea of permutation matrix $\mathcal{P}$ to the aforementioned geometric transformations. 

In the view of invariant feature learning, the permutation invariance of $\mathcal{F}$ outputs are guaranteed through the local aggregations of node features of $x \in X$ and the linear transformations of aggregated node features:
\begin{displaymath}
\label{eq:perm_f}
\begin{aligned}
\mathcal{F}(X;\theta) = \rho(\sum_{x \in X} \phi(x, \theta)). 
\end{aligned}
\end{displaymath}
The local differentiable function $\phi$ linearly transforms each node feature individually $x$ to latent space and the aggregation function sums over latent node features \footnote{The basic aggregation in this setting can also be averaging or max pooling.}, and $\rho$ is a global differentiable function applied on the summed node features followed by a non-linear activation function. In this simple setting, the outputs of $\mathcal{F}$ are invariant to the permutation of node features since the summation-based aggregation returns the same outputs for any input permutation: $Aggr_{sum}(x_1, x_2,....,x_n) = Aggr_{sum}(x_n,...,x_2, x_1)$. 

This framework (i.e., DeepSet \citep{zaheer2017deep}) assumes that permutations affect only individual node features while disregarding their connectivity, which is crucial for polygonal structures. Graph convolution neural networks (GCN) by \citet{kipf2017semi} learn convolutional features of node matrix $X$ with the normalized graph Laplacian matrix $\hat{A} = \tilde{D}^{-1/2}\tilde{A}\tilde{D}^{-1/2}$, where $\tilde{A} = A + I$ is the adjacency matrix with a identity matrix (i.e., self-connections of nodes), and $\tilde{D}$ is the diagonal degree matrix. The graph Laplacian matrix describes the divergence of energy flowing from source nodes to target nodes, analogous to the spatial relationships in polygonal structures. Specifically, it captures connectivity patterns both within polygons (e.g., rings with holes) and between adjacent polygons.

Thus, GCN can learn graph Laplacian embedding of node feature with a differentiable neural network layer as follows: 
\begin{equation} 
\label{eq:matrix_gcn}
\begin{array}{c}
\mathcal{F}_{conv}(X, A; \theta) = \tilde{D}^{-1/2}\tilde{A}\tilde{D}^{-1/2}X\theta\\  
= \hat{A}X\theta.
\end{array}
\end{equation}
We can express Eq.\ref{eq:matrix_gcn} from a vector-wise view:
\begin{equation} 
\label{eq:node_gcn}
f(x_i; a_{ij}, \theta) = \sum_{\substack{a_{ij} \in A, \\ x \in X}} \phi(x_i, x^{\prime}_j, \theta), \text{where} \; x^{\prime}_j = \frac{\tilde{a}_{ij}}{\sqrt{\tilde{d}_{i} \times \tilde{d}_j}}x_j,
\end{equation}
$\frac{\tilde{a}_{ij}}{\sqrt{\tilde{d}_{i} \times \tilde{d}_j}}$ corresponds to the edge weight between nodes $x_i$ and $x_j$ in the normalized graph Laplacian matrix $\hat{A}$. From Eq.\ref{eq:node_gcn}, the latent feature of target node $x_i$ depends on the relation with its neighbour node $x_j$ and constant edge weight $\frac{\tilde{a}_{ij}}{\sqrt{\tilde{d}_{i} \times \tilde{d}_j}}$. However, the constant edge weight between nodes $x_i$ and $x_j$ largely limits the expressivity of latent node features of polygonal graph since every node has a constant node degree ($d=2$). 

As discussed in \citet{yan2021graph}, graph autoencoders GCAE based on GCN are sensitive to orientation and rotation of polygons, and therefore current graph-based learning methods are not optimized for geospatial settings with polygonal geometries of arbitrary orientation, or conversely, arbitrary observer orientation. Here we propose to leverage the message-passing mechanism \citep{gilmer2017neural,battaglia2018relational} with the permutation-invariant function $\mathcal{F}$ to learn expressive and robust latent features of polygons. 

\section{Approach}

\subsection{Graph Representation of Polygon}
\label{sec:graph_rep_poly}
Polygons are sets of points connected by lines, forming a collection of clockwise or counterclockwise linear rings \citep{open2003opengis}. According to the Gestalt principles of invariance, the compact embedding of linear rings should ideally be invariant to geometric transformations such as rotation, scaling, and shear. We define a graph $\mathcal{G} = (X, E)$, where $X$ is a node matrix containing the coordinates of the geometry vertices, and $E$ is an edge matrix capturing the connectivity of these vertices along the linear rings (i.e., the boundaries of the polygon and any holes). The adjacency matrix $A$ encodes the connectivity between nodes, where a binary value $a_{ij} = 1$ indicates that nodes $i$ and $j$ are connected.

\subsection{Message-passing Encoder}
\label{sec:mp_nn}
The principle of message-passing involves aggregating neighboring node features in graphs based on messages computed between source and target nodes. We define the message-passing function operating on a single node as:
\begin{equation} 
\label{eq:message-passing}
\begin{array}{c}
h_i = \phi(x_i, msg(x_i, x_j), \theta), \;
\end{array}
\end{equation}
where $\{ x_i, x_j\} \in A = 1$ and $msg(\cdot)$ is the message function that takes the source and target nodes as inputs. From Eq.\ref{eq:message-passing}, the node feature of $x_i$ is updated with the computed message $(x_i, x_j)$ and transformed linearly into latent space, producing the latent feature $h_i \in \mathcal{H}$. The permutation-invariant message-passing encoder is represented in vector form as:
\begin{equation} 
\label{eq:mpn}
\begin{array}{c}
\mathcal{F}_{msg}(X, A;\theta) = \rho_{\theta}(readout(h_i | h_i \in \mathcal{H})).
\end{array}
\end{equation},
where a global readout function (e.g., mean or max pooling) aggregates the latent node-wise features ($\mathcal{H}$) and computes a global graph embedding by applying a non-linear transformation to the pooled features via $\rho_{\theta}$.

Building on geometric message-passing methods for point clouds \citep{wang2019dynamic,qi2017pointnet++} and polygons \citep{liu2021triangleconv}, we define a specific message-passing function $msg(\cdot)$ for PolyMP as:
\begin{equation} 
\label{eq:message-function}
\begin{array}{c}
msg(x_i, x_j) = \alpha * \dfrac{(x_j - x_i)}{std_{\mathcal{N}_{x_i}} + \epsilon} + \beta.
\end{array}
\end{equation}, where $\alpha$ and $\beta$ are learnable parameters that adjust the local neighborhoods to a normal distribution, and $std_{\mathcal{N}_{x_i}}$ computes the standard deviation of the local neighborhoods $\mathcal{N}{x_i}$ with a small $\epsilon$ to ensure numerical stability. This operation facilitates stable geometric feature learning of graph neighborhoods across diverse geometric structures \citep{marethinking}, accommodating the varying properties of polygons (Fig.\ref{fig:glyph_trans}). After the message-passing encoder layers, a global pooling layer aggregates the node features to generate a whole-graph embedding, which is then used for downstream polygon shape classification.

\begin{figure}[htbp]
\centering
\includegraphics[width=1\columnwidth]{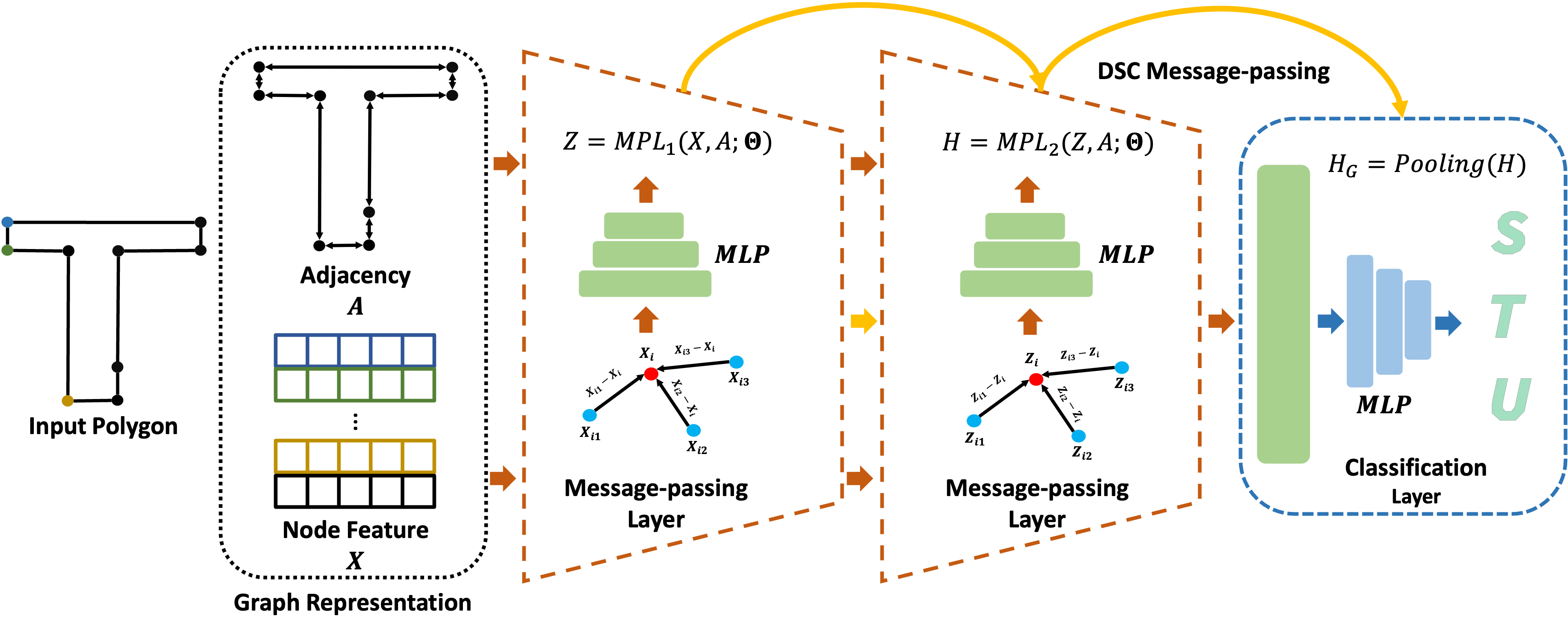}
\caption{Model architecture of PolyMP and PolyMP-DSC. PolyMP consists of a simple message-passing encoder followed by a classification layer (multi-layer perceptron). Yellow arrows indicate the densely self-connected message-passing in PolyMP-DSC.}
\label{fig:mpn}
\end{figure}

Following Eq. \ref{eq:message-passing}-\ref{eq:message-function}, we introduce a simple \textbf{M}essage-\textbf{p}assing neural network for \textbf{Poly}gon geometries, named PolyMP, as shown in Fig. \ref{fig:mpn}. PolyMP processes polygon point coordinates as input node features and the connectivity of polygon edges as input edges. The message-passing encoders of PolyMP compute the \enquote{message} for each node based on the graph adjacency and update the node features in each layer.

\subsection{Densely Self-Connected Message-Passing}
\label{sec:densely_connected_message_passing}
As noted in recent studies of graph message-passing networks \citep{fan2021propagation}, the basic message-passing mechanism restricts information flow due to the absence of self-loop information in graph representation learning. To address this limitation and enhance PolyMP's learning capability, we adopt a densely self-connected (DSC) message-passing mechanism \citep{huang2017densely,fan2021propagation,he2016deep}, defined as:

\begin{equation} 
\label{eq:densely_self_connected} 
\begin{array}{c} 
\mathcal{G}_l = f(\rho_{\theta}(readout(h_i | h_i \in \mathcal{H}_{l})) + \mathcal{G}_{l-1} + \mathcal{G}_{l-2} + ... + \mathcal{G}_{0}), 
\end{array} 
\end{equation} where we extend the basic PolyMP model to PolyMP-DSC, as shown in Fig. \ref{fig:mpn}. The PolyMP-DSC aggregates latent graph-wise features $\mathcal{G}_l$ at layer $l$ by incorporating latent graph-wise features $\mathcal{G}_{l-1...0}$ from previous message-passing layers. This enables the model to capture hierarchical self-looped graph information, resulting in richer graph-wise representations.

\subsection{Implementation}
\label{implementation}

\begin{figure}[htbp]
\centering
\includegraphics[width=1\columnwidth]{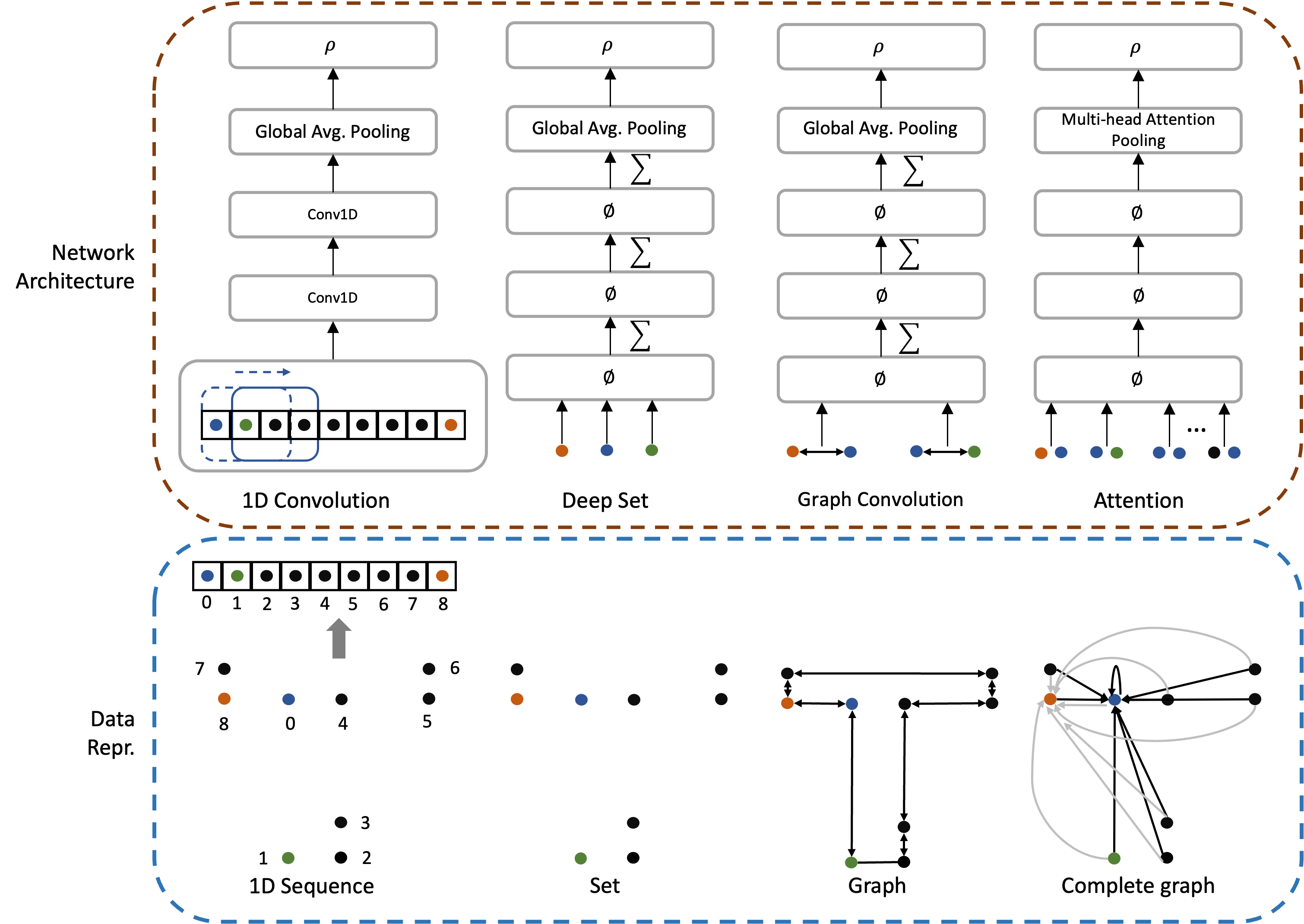}
\caption{Data representations of polygons (as 1D sequence, Point Set, Graph and Complete Graph) and the corresponding learning models (Baselines: VeerCNN, DeepSet, GCAE and SetTransformer).}
\label{fig:model_config}
\end{figure}

\begin{table}[htbp]
\centering
\caption{Model configuration. Each model is configured to have the same number of convolution layers and dimensions of latent embedding, to keep the model size (i.e., number of training parameters) comparable.}
\begin{tabular}{lccrc}
\toprule
\textbf{Model} & \parbox[t]{0.9cm}{\textbf{Conv.\\depth}} & \parbox[t]{0.9cm}{\textbf{Latent\\dim.}} & \textbf{\#Param.} & \textbf{Pooling}\\
\midrule
\shortstack{VeerCNN \citep{veer2018deep}} & 2 & \{2, 32, 64\} & 13,853 & Global avg.\\
\shortstack{DeepSet \citep{zaheer2017deep}} & 2 & \{2, 64, 64\} & 11,837 & Global avg.\\
\shortstack{SetTransformer \citep{lee2019set}} & 2 & \{2, 64, 64\} & 113,531 & Attention\\
\shortstack{GCAE \citep{yan2021graph}} & 2 & \{2, 64, 64\} & 7,613 & Global avg.\\
\shortstack{NUFTSpec \citep{mai2023towards}} & 2 & \{288, 128, 64\} & 48,635 & - \\
\shortstack{DSC-NMP \citep{fan2021propagation}} & 2 & \{2, 64, 64\} & 20,539 & Global add. \\
\rowcolor{lightgray} \shortstack{PolyMP (Ours)} & 2 & \{2, 64, 64\} & 11,837 & Global max\\
\rowcolor{lightgray} \shortstack{PolyMP-DSC (Ours)} & 2 & \{2, 64, 64\} & 16,579 & Global max\\
\bottomrule
\end{tabular}
\label{tab:model_config}
\end{table}

We evaluate the performance of learning models from previous studies, specifically: the VeerCNN model from \citet{veer2018deep}, which uses 1D sequences converted from polygons as input; the DeepSet model from \citet{zaheer2017deep} and SetTransformer model from \citet{lee2019set}, which treat polygon vertices as input point clouds; the GCAE model from \citet{yan2021graph}, which represents polygons as input graphs; and the DSC-NMP model from \citet{fan2021propagation}, a baseline graph learning model that processes polygons as input graphs and applies DSC neural message-passing for graph feature learning. We also evaluate the NUFTSpec model from \citet{mai2023towards}, which encodes polygon geometries by converting them into feature vectors in the rasterized spectral domain. These models are compared against the proposed graph message-passing models for polygons, \textit{PolyMP} and \textit{PolyMP-DSC}.

The model configurations used in our experiments are summarized in Table~\ref{tab:model_config}, and the data representations of polygons and baseline learning models are illustrated in Figure \ref{fig:model_config}. For experimental control, every model trained and tested was constrained to a comparable model depth and latent space of feature embedding. During model training on the Glyph dataset and fine-tuning on the OSM dataset, we compute cross-entropy loss \citep{zhang2018generalized} to determine training gradients for fast convergence and good model generalizability. We use Adam optimization \citep{DBLP:journals/corr/KingmaB14} with an initial learning rate of 0.01. For better model generalization, we adopt a learning rate decay strategy that reduces the learning rate by a factor of 10 once the model training reaches a performance plateau after 25 epochs. An early-stopping mechanism halts the training once the loss ceases to decrease for 50 epochs, mitigating overfitting. The batch size is set to 64, and the total training epochs are set to 100. Importantly, the models are not designed to individually outperform state-of-the-art models for each architecture, but rather to maintain comparability across architectures and enable the evaluation of the effects of data representations on results.

\section{Experiment}
\label{experiment}

\subsection{Glyph Dataset}
\label{glyph}
We introduce a synthetic dataset of highly variable geometric shapes to benchmark the classification performance of learning models on vector polygons. This dataset consists of 26 Latin alphabet character glyph geometries, representing semantic classes from \textit{A} to \textit{Z}, gathered from an online source \citep{google_2010}. A similar approach has been used previously in computational geometry to construct rich datasets for algorithm testing \citep{duckham2008efficient}.

The synthetic dataset includes the boundaries (contour lines) of glyphs extracted from 1,413 sans serif and 1,002 serif fonts, resulting in 2D simple polygon geometries compliant with the standards of \citet{open2003opengis}. Serif and sans serif fonts are the two primary typographic families. Serif glyphs feature decorative strokes that enhance the legibility of body text, while sans serif glyphs have clean, minimal strokes, making them more suitable for headers. Importantly, these minor variations do not affect the overall Gestalt of the shapes, as readers can consistently recognize the symbols. This stability ensures that the labels assigned to the symbols are both stable and robust.

\begin{figure}[htbp]
\centering
\includegraphics[width=1\columnwidth]{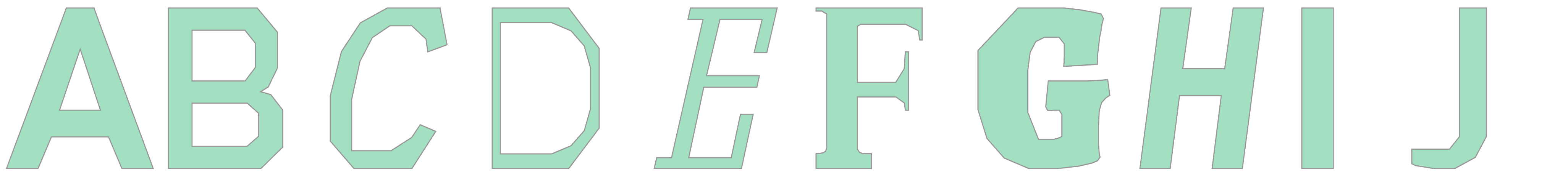}
\caption{Glyph dataset samples. Left to right columns: A-shape to J-shape glyphs.}
\label{fig:glyph}
\end{figure}

Each polygon geometry is encoded into a fixed-size feature matrix of size $\in \mathbb{R}^{n \times 3}$, where each feature vector $\in \mathbb{R}^3$ contains the 2D coordinates of vertices $(x, y)$ as the geometric feature, and a binary feature $(0, 1)$ to indicate the position of each 2D coordinate (whether it lies on the outer rings or holes of the polygons). Examples of glyph geometries are shown in Figure \ref{fig:glyph}.

\subsection{OSM Dataset}
\label{osm}
\begin{figure}[htbp]
\centering
\includegraphics[width=1\columnwidth]{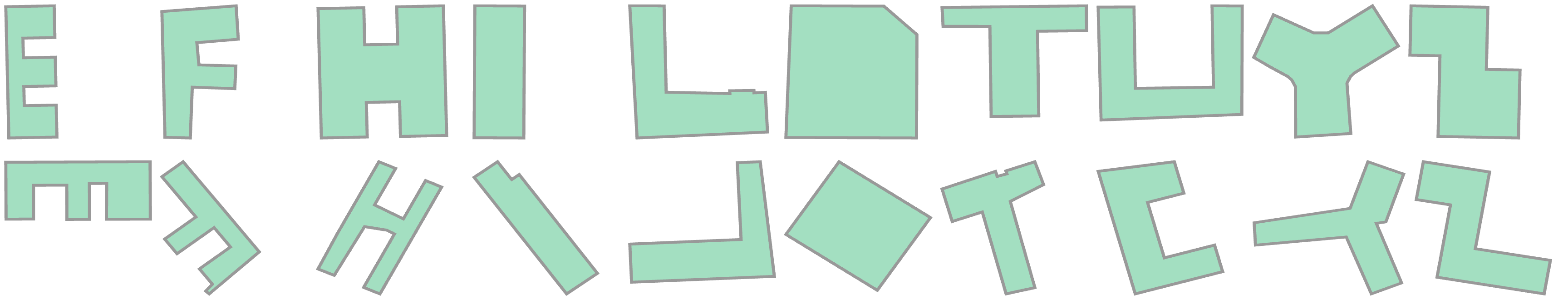}
\caption{OSM dataset samples. Top row: building geometries of standard shapes. Bottom row: geometric transformation counterparts. Left to right columns: E, F, H, I, L, O, T, U, Y and Z-shape buildings. \citep{yan2021graph}}
\label{fig:osm}
\end{figure}

Latin alphabet glyph geometries share geometric similarities with building footprints. To test the generalizability of learning models on spatial objects, we use a building geometry dataset proposed by \citet{yan2021graph}. The dataset contains 10,000 real-world building footprints extracted from OpenStreetMap (OSM) \citep{OpenStreetMap}, labeled into 10 categories based on template matching to letters \citep{yan2017template}. The building footprints are randomly rotated and reflected from the original canonical samples, as shown in Figure~\ref{fig:osm}. This dataset is currently the most comprehensive real-world collection of vector geometries with purely shape-based labels (as opposed to labels based on building use, which should not be used for shape classification). However, we note that the assignment of labels in this dataset is somewhat subjective and may be less robust compared to the glyph-based dataset. Specifically, shape reflection can pose challenges for label stability, especially for asymmetric letters (e.g., $R$) or mirror image shapes of distinct labels (e.g., $S$ and $Z$).

\subsection{Pre-processing}
\begin{figure}[htbp]
\centering
\includegraphics[width=1\columnwidth]{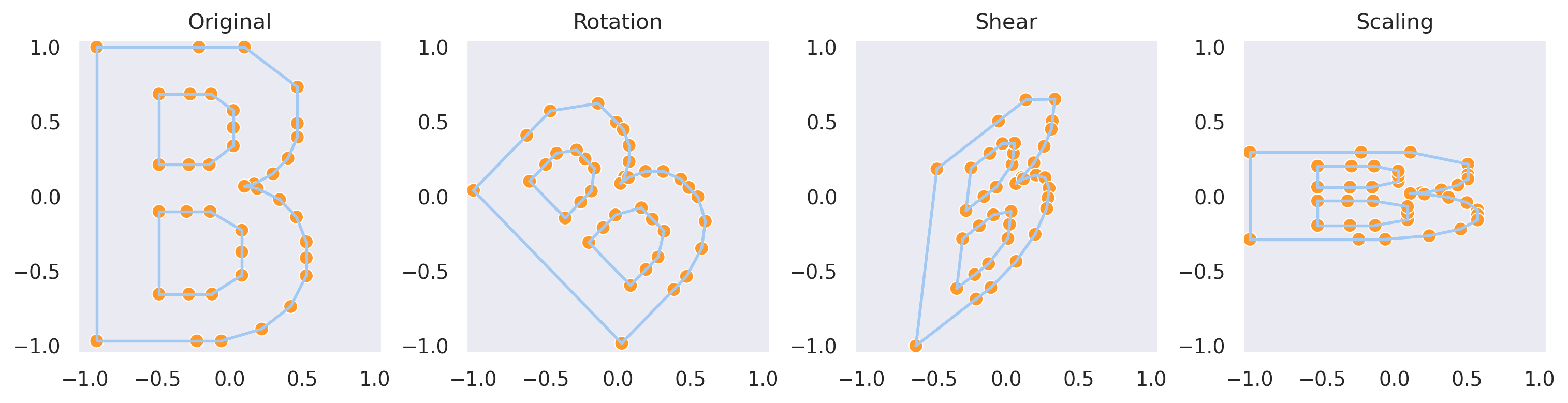}
\caption{Data samples of Glyph dataset under label-preserving transformations.}
\label{fig:glyph_trans}
\end{figure}

\paragraph{Label-preserving transformations} To assess the invariance of geometric transformations in polygon representations and their effect on learning models, we apply label-preserving geometric transformations to the glyph polygons. We create four transformed training datasets, each consisting of 20\%, 40\%, 60\%, and 80\% of glyph polygons that have undergone these transformations. The details of the transformations are listed in Table \ref{tab:glyph_trans}, with the original and transformed data samples visualized in Fig.~\ref{fig:glyph_trans}. This results in five training datasets: 0\%, 20\%, 40\%, 60\%, and 80\% transformations, which will be used for the experiment. The test set consists of data randomly sampled from these five datasets.

\begin{table}[htbp]
\centering
\caption{Label-preserving transformations applied to samples of Glyph dataset maintains the semantic information (label) of original polygons (i.e., 180 degrees rotation of letter M alter its semantic label, converting to letter W).}
\begin{tabular}{ll}
\toprule
\textbf{Transformation} & \textbf{Operation} \\
\midrule
Rotation & Rotate data around its centroid by a random angle $\in$ $\left[ -75^{\circ},75^{\circ}\right]$.\\ 
Scaling & Scale data by random distinct factors $\in$ $\left[ 0.1,2\right]$ on the $x$ and $y$ axes.\\ 
Shearing & Shear data in a random angle $\in$ $\left[ -45^{\circ},45^{\circ}\right]$ on the $x$ and $y$ axes.\\
\bottomrule
\end{tabular}
\label{tab:glyph_trans}
\end{table}

\begin{figure}[htbp]
\centering
\includegraphics[width=1\columnwidth]{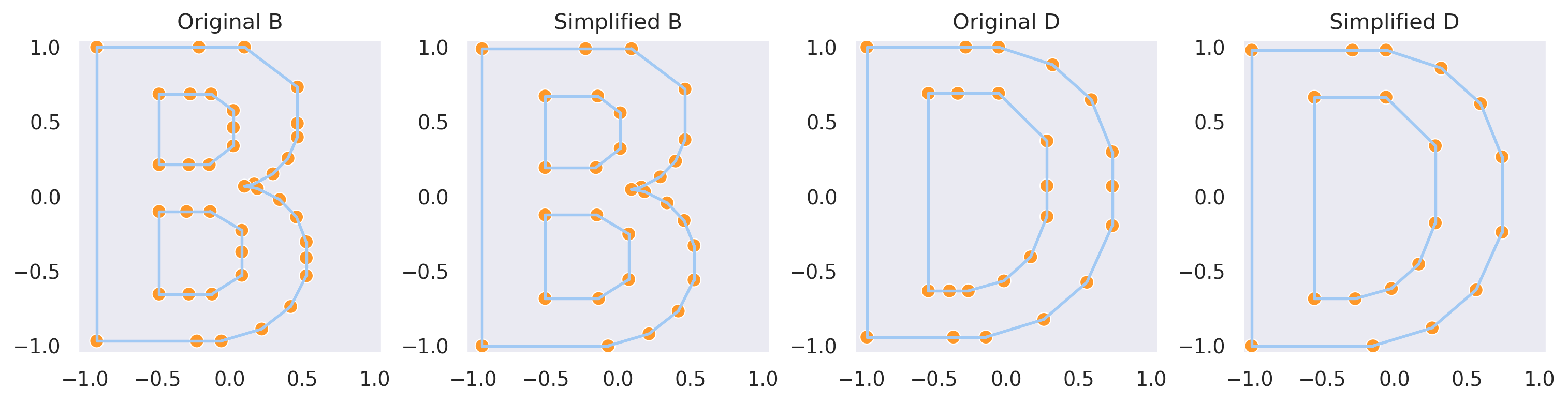}
\caption{Visualisation of original and simplified polygon examples generated with Douglas–Peucker algorithm \citep{douglas1973algorithms}. The simplified polygons, compared to original polygons, contain a subset of the vertices that defined the original curve.}
\label{fig:glyph_simp}
\end{figure}

\paragraph{Polygon Simplifications} To assess the invariance of polygon representations to trivial vertices, we apply the Douglas–Peucker algorithm \citep{douglas1973algorithms} with a tolerance of 1.0 to the original glyph polygons (typically sized around 50 $\times$ 50 units). This simplification process generates polygon samples with fewer vertices, particularly by excluding co-linear or nearly co-linear vertices. Importantly, the simplification is applied before normalizing the polygons to the range of (-1, 1) and before any augmentation, ensuring that the simplified geometries preserve the topology and essential shape of the original glyph (see Fig.~\ref{fig:glyph_simp}).

\section{Result}
\label{results}
We report model performances for all experiments as the overall accuracy (O.A.), calculated as the ratio of true positives over the total number of samples. 

\subsection{Results on Glyph Dataset}

\begin{table}[htbp]
\small
\centering
\caption{Test performances on Glyph dataset. Transformation ratios 0\%, 20\%, 40\%, 60\% and 80\% refer to training datasets consisting of 0\%, 20\%, 40\%, 60\% and 80\% geometric transformed samples. Glyph-O, Glyph-R, Glyph-SC and Glyph-SH Acc. indicate the classification accuracy of models tested on original, rotated, scaled and sheared data. Overall accuracy is reported as Glyph O.A..}
\begin{tabular}{lcccccc}
\toprule
\textbf{\shortstack{Trans. \\ Ratio}} & \textbf{Model} & \textbf{\shortstack{Glyph-O \\ Acc.}} & \textbf{\shortstack{Glyph-R \\ Acc.}} & \textbf{\shortstack{Glyph-SC \\ Acc.}} & \textbf{\shortstack{Glyph-SH \\ Acc.}} & \textbf{\shortstack{Glyph \\ O.A.}} \\
\midrule
\multirow{8}*{0\%} 
& \shortstack{VeerCNN} & 97.64 & 41.51 & 79.15 & \textbf{59.07} & 69.43 \\
& \shortstack{DeepSet} & 94.62 & 23.67 & 65.01 & 26.00 & 52.56 \\
& \shortstack{SetTransformer} & 96.25 & 25.23 & 67.91 & 27.79 & 54.42 \\
& \shortstack{GCAE} & 95.16 & 31.38 & 64.54 & 38.84 & 57.83 \\
& \shortstack{NUFTSpec} & 98.42 & 34.90 & 72.07 & 39.68 & 61.34 \\
& \shortstack{DSC-NMP} & 96.81 & 35.71 & 72.83 & 43.21 & 62.24 \\
\rowcolor{lightgray} & \shortstack{PolyMP} & \textbf{99.68} & 39.91 & 82.35 & 53.22 & 69.03 \\
\rowcolor{lightgray} & \shortstack{PolyMP-DSC} & 99.58 & \textbf{41.74} & \textbf{86.45} & 56.16 & \textbf{71.21} \\
\midrule
\multirow{8}*{20\%} 
& \shortstack{VeerCNN} & 95.12 & 67.63 & 83.53 & 77.78 & 81.05 \\
& \shortstack{DeepSet} & 93.88 & 67.95 & 77.31 & 64.72 & 76.25 \\
& \shortstack{SetTransformer} & 93.87 & 71.72 & 82.43 & 72.56 & 80.18 \\
& \shortstack{GCAE} & 93.31 & 47.31 & 71.47 & 57.11 & 67.61 \\
& \shortstack{NUFTSpec} & 98.00 & 65.27 & 84.03 & 70.74 & 79.61 \\
& \shortstack{DSC-NMP} & 93.68 & 52.38 & 78.86 & 61.72 & 71.73 \\
\rowcolor{lightgray} & \shortstack{PolyMP} & 99.10 & 84.89 & 93.26 & 88.17 & 91.30 \\
\rowcolor{lightgray} & \shortstack{PolyMP-DSC} & \textbf{99.19} & \textbf{86.20} & \textbf{94.60} & \textbf{90.65} & \textbf{92.64} \\
\midrule
\multirow{8}*{40\%} 
& \shortstack{VeerCNN} & 94.75 & 73.26 & 84.77 & 80.29 & 83.30 \\
& \shortstack{DeepSet} & 92.57 & 74.24 & 79.24 & 72.08 & 80.12 \\
& \shortstack{SetTransformer} & 91.97 & 72.73 & 83.25 & 73.25 & 80.33 \\
& \shortstack{GCAE} & 91.70 & 59.55 & 74.35 & 64.42 & 72.72 \\
& \shortstack{NUFTSpec} & 97.44 & 78.84 & 86.00 & 78.71 & 85.24 \\
& \shortstack{DSC-NMP} & 86.71 & 54.01 & 71.96 & 57.97 & 67.71 \\
\rowcolor{lightgray} & \shortstack{PolyMP} & 98.85 & 88.53 & 94.35 & 90.74 & 93.11 \\
\rowcolor{lightgray} & \shortstack{PolyMP-DSC} & \textbf{99.05} & \textbf{90.18} & \textbf{95.33} & \textbf{92.57} & \textbf{94.18} \\
\midrule
\multirow{8}*{60\%} 
& \shortstack{VeerCNN} & 94.50 & 77.16 & 85.38 & 82.36 & 84.88 \\
& \shortstack{DeepSet} & 91.72 & 76.89 & 81.46 & 73.97 & 81.49 \\
& \shortstack{SetTransformer} & 90.93 & 77.04 & 82.39 & 77.31 & 81.94 \\
& \shortstack{GCAE} & 90.46 & 65.02 & 76.38 & 68.78 & 75.50 \\
& \shortstack{NUFTSpec} & 96.80 & 83.13 & 88.40 & 81.76 & 87.55 \\
& \shortstack{DSC-NMP} & 90.97 & 64.17 & 80.00 & 67.73 & 75.76 \\
\rowcolor{lightgray} & \shortstack{PolyMP} & 98.59 & 90.08 & 94.11 & 91.65 & 93.69 \\
\rowcolor{lightgray} & \shortstack{PolyMP-DSC} & \textbf{98.79} & \textbf{91.80} & \textbf{95.76} & \textbf{93.20} & \textbf{94.91} \\
\midrule
\multirow{8}*{80\%} 
& \shortstack{VeerCNN} & 93.51 & 78.97 & 85.25 & 83.17 & 85.25\\
& \shortstack{DeepSet} & 90.66 & 78.68 & 82.14 & 76.55 & 82.47 \\
& \shortstack{SetTransformer} & 91.17 & 79.94 & 83.69 & 79.78 & 83.67 \\
& \shortstack{GCAE} & 89.58 & 67.95 & 77.99 & 70.55 & 76.83 \\
& \shortstack{NUFTSpec} & 96.15 & 84.40 & 89.24 & 83.46 & 88.32 \\
& \shortstack{DSC-NMP} & 94.31 & 74.31 & 86.75 & 78.36 & 83.46 \\
\rowcolor{lightgray} & \shortstack{PolyMP} & 98.48 & 91.64 & 94.69 & 92.36 & 94.31 \\
\rowcolor{lightgray} & \shortstack{PolyMP-DSC} & \textbf{98.93} & \textbf{93.19} & \textbf{95.95} & \textbf{94.26} & \textbf{95.61} \\
\bottomrule
\end{tabular}
\label{tab:glyph}
\end{table}

\begin{table}[htbp]
\small
\centering
\caption{Test performances on Glyph dataset, consisting of simplified polygons generated by Douglas–Peucker algorithm (tolerant=1.0) with the removal of trivial vertices.}
\begin{tabular}{lcccccc}
\toprule
\textbf{\shortstack{Trans. \\ Ratio}} & \textbf{Model} & \textbf{\shortstack{Glyph-O \\ Acc.}} & \textbf{\shortstack{Glyph-R \\ Acc.}} & \textbf{\shortstack{Glyph-SC \\ Acc.}} & \textbf{\shortstack{Glyph-SH \\ Acc.}} & \textbf{\shortstack{Glyph \\ O.A.}} \\
\midrule
\multirow{8}*{0\%}
& \shortstack{VeerCNN} & 94.72 & 39.65 & 75.42 & 56.12 & 66.56 \\
& \shortstack{DeepSet} & 94.01 & 23.75 & 65.11 & 25.53 & 52.30 \\
& \shortstack{SetTransformer} & 94.57 & 23.67 & 65.72 & 26.61 & 52.76 \\
& \shortstack{GCAE} & 94.83 & 31.64 & 64.21 & 39.90 & 58.08 \\
& \shortstack{NUFTSpec} & 98.03 & 34.91 & 71.96 & 40.07 & 61.30 \\
& \shortstack{DSC-NMP} & 95.40 & 35.21 & 71.61 & 41.56 & 61.04 \\
\rowcolor{lightgray} & \shortstack{PolyMP} & 99.27 & 40.52 & 81.90 & 53.50 & 68.92 \\
\rowcolor{lightgray} & \shortstack{PolyMP-DSC} & \textbf{99.36} & \textbf{42.26} & \textbf{86.16} & \textbf{56.68} & \textbf{71.07} \\
\midrule
\multirow{8}*{20\%} 
& \shortstack{VeerCNN} & 90.55  & 62.58 & 77.80 & 72.31 & 75.85 \\
& \shortstack{DeepSet} & 93.81 & 68.26 & 77.83 & 64.93 & 76.47 \\
& \shortstack{SetTransformer} & 92.31 & 70.51 & 80.54 & 70.65 & 78.54\\
& \shortstack{GCAE} & 92.26 & 47.29 & 71.28 & 57.33 & 67.39 \\
& \shortstack{NUFTSpec} & 97.76 & 65.10 & 83.97 & 71.41 & 79.65 \\
& \shortstack{DSC-NMP} & 92.20 & 51.73 & 77.60 & 60.44 & 70.55 \\
\rowcolor{lightgray} & \shortstack{PolyMP} & 98.37 & 83.37 & 91.41 & 87.25 & 89.74 \\
\rowcolor{lightgray} & \shortstack{PolyMP-DSC} & \textbf{98.93} & \textbf{84.57} & \textbf{93.56} & \textbf{89.15} & \textbf{91.19} \\
\midrule
\multirow{8}*{40\%} 
& \shortstack{VeerCNN} & 89.63 & 67.82 & 78.72 & 74.61 & 77.73 \\
& \shortstack{DeepSet} & 92.66 & 74.81 & 79.60 & 71.62 & 80.46 \\
& \shortstack{SetTransformer} & 89.86 & 70.91 & 80.70 & 70.97 & 78.14 \\
& \shortstack{GCAE} & 91.65 & 59.30 & 74.95 & 64.79 & 72.94 \\
& \shortstack{NUFTSpec} & 97.23 & 78.85 & 86.09 & 78.95 & 85.25 \\
& \shortstack{DSC-NMP} & 85.08 & 52.92 & 69.90 & 56.73 & 66.21 \\
\rowcolor{lightgray} & \shortstack{PolyMP} & 98.10 & 86.47 & 92.72 & 88.73 & 91.09 \\
\rowcolor{lightgray} & \shortstack{PolyMP-DSC} & \textbf{98.66} & \textbf{89.04} & \textbf{94.45} & \textbf{91.84} & \textbf{93.28} \\
\midrule
\multirow{8}*{60\%} 
& \shortstack{VeerCNN} & 89.29 & 71.91 & 79.42 & 77.29 & 79.50 \\
& \shortstack{DeepSet} & 91.58 & 77.44 & 81.62 & 73.84 & 81.57 \\
& \shortstack{SetTransformer} & 89.70 & 75.03 & 80.26 & 75.53 & 80.15 \\
& \shortstack{GCAE} & 90.06 & 63.61 & 76.64 & 68.05 & 74.79 \\
& \shortstack{NUFTSpec} & 96.62 & 82.49 & 88.48 & 82.16 & 87.46 \\
& \shortstack{DSC-NMP} & 90.02 & 62.99 & 79.47 & 66.67 & 74.83 \\
\rowcolor{lightgray} & \shortstack{PolyMP} & 97.74 & 89.50 & 92.99 & 90.59 & 92.48 \\
\rowcolor{lightgray} & \shortstack{PolyMP-DSC} & \textbf{98.38} & \textbf{91.53} & \textbf{94.89} & \textbf{92.76} & \textbf{94.13} \\
\midrule
\multirow{8}*{80\%} 
& \shortstack{VeerCNN} & 88.05 & 73.89 & 79.55 & 77.17 & 79.69 \\
& \shortstack{DeepSet} & 91.04 & 79.21 & 82.44 & 76.45 & 82.67 \\
& \shortstack{SetTransformer} & 89.99 & 79.14 & 82.70 & 78.60 & 82.63\\
& \shortstack{GCAE} & 88.22 & 66.55 & 78.01 & 70.44 & 76.06 \\
& \shortstack{NUFTSpec} & 96.10 & 84.56 & 89.16 & 83.28 & 88.27 \\
& \shortstack{DSC-NMP} & 93.06 & 73.10 & 85.03 & 76.99 & 82.08 \\
\rowcolor{lightgray} & \shortstack{PolyMP} & 97.42 & 90.55 & 93.25 & 91.36 & 92.97 \\
\rowcolor{lightgray} & \shortstack{PolyMP-DSC} & \textbf{98.47} & \textbf{92.78} & \textbf{95.10} & \textbf{93.43} & \textbf{94.77} \\
\bottomrule
\end{tabular}
\label{tab:glyph_sim_1.0}
\end{table}

Table~\ref{tab:glyph} and Table~\ref{tab:glyph_sim_1.0} present the model performances on the Glyph dataset and simplified polygons, respectively.

At the $0\%$ transformation ratio in Table~\ref{tab:glyph}, we observe that PolyMP achieves $99.68\%$, $39.91\%$, and $82.35\%$ accuracy on the original, rotated, and scaled test samples, respectively, while VeerCNN achieves $59.07\%$ accuracy on sheared samples. In comparison, DeepSet, SetTransformer, and GCAE maintain comparable performance on the original and scaled test samples ($94.62\%$, $65.01\%$; $96.25\%$, $67.91\%$; and $95.16\%$, $64.54\%$, respectively). However, models that learn from set (DeepSet and SetTransformer) and graph inputs (GCAE and DSC-NMP) experience significant performance deterioration on rotated and sheared samples, with accuracy of $23.67\%$, $26.00\%$; $25.23\%$, $27.79\%$; $31.38\%$, $38.84\%$; and $35.71\%$, $43.21\%$, respectively. Overall, PolyMP-DSC, which learns densely connected graph features, achieves the highest classification accuracy at $71.21\%$, significantly outperforming the baseline DSC-NMP model ($62.24\%$ O.A.).

At the 20\% transformation ratio, both PolyMP (84.89\% (+44.98\%), 93.26\% (+10.91\%), and 88.17\% (+34.95\%)) and PolyMP-DSC (86.20\% (+44.46\%), 94.60\% (+8.15\%), and 90.65\% (+34.49\%)) show significant improvements in test accuracy on rotated, scaled, and sheared samples, respectively. In terms of overall accuracy, PolyMP achieves 91.30\% O.A. (+22.27\%), outperforming VeerCNN's 81.05\% O.A. PolyMP-DSC records the highest O.A. at 92.64\% (+21.43\%). In contrast, the baseline learning models are unable to match the performance of PolyMP and PolyMP-DSC, despite the increase in training set diversity.

At the 40\% and 60\% transformation ratios, we observe that the accuracy improvements from increasing the proportion of geometrically transformed samples in the training set have plateaued. In this case, PolyMP achieves 93.11\% and 93.69\% O.A., while PolyMP-DSC leads with 94.18\% and 94.91\% O.A. Both models maintain their superior performance compared to others. VeerCNN records 83.30\% and 84.88\% O.A., followed by DeepSet (80.12\% and 81.49\% O.A.), SetTransformer (80.33\% and 81.94\% O.A.), GCAE (72.72\% and 75.50\% O.A.), DSC-NMP (57.97\% and 67.73\% O.A.), and NUFTSpec (70.74\% and 78.71\% O.A.), all of which fall behind.

At the 80\% transformation ratio, PolyMP and PolyMP-DSC achieve the highest test performances on rotated and sheared samples, with accuracy of 91.64\%, 92.36\%, and 93.19\%, 94.26\%, respectively. These empirical results suggest that, compared to sequence encoding of polygons using CNNs and set representations with set-based learning models, graph representations combined with message-passing neural networks (PolyMP and PolyMP-DSC) are more robust to permutations of polygon vertices. This robustness is particularly evident in geometric transformations such as rotation and shearing, where the feed-in order of vertices can vary. Comparing basic PolyMP with the extended PolyMP-DSC, we observe testing performance improvements of up to $2\%$ O.A. on the Glyph dataset. These results highlight the effectiveness of incorporating hierarchical self-looped graph representations via densely self-connected message passing for polygon encoding and shape-based classification.

From a different perspective, as the proportion of original samples progressively decreases in the training set (from 0\% to 80\% transformation ratios), PolyMP and PolyMP-DSC experience only a slight test performance deterioration, with a decrease from 99.68\% to 98.48\% (-1.20\%) and 99.58\% to 98.93\% (-0.65\%), respectively. In contrast, VeerCNN (-4.13\%), DeepSet (-3.96\%), SetTransformer (-5.08\%), GCAE (-5.58\%), NUFTSpec (-2.27\%), and DSC-NMP (-2.50\%) exhibit more significant performance drops in test accuracy on the original samples.

In Table~\ref{tab:glyph_sim_1.0}, we present the test performances of models trained and evaluated on simplified polygons from the Glyph dataset. This experimental setting focuses on evaluating the models' performance on polygons with a reduced number of trivial vertices (Fig. \ref{fig:glyph_simp}) and compares these results with those shown in Table~\ref{tab:glyph}.

We observe similar performance trends in the models when evaluated on the simplified polygons. Notably, VeerCNN experiences a significant drop in performance compared to the results in Table~\ref{tab:glyph}. Specifically, at the 0\% transformation ratio, VeerCNN's test performance decreases from 69.43\% O.A. (Table~\ref{tab:glyph}) to 66.56\% O.A. (Table~\ref{tab:glyph_sim_1.0}), representing a 2.87\% decline. As data diversity increases, the performance deterioration in VeerCNN becomes more pronounced, with a drop of 5.2\%, progressively increasing to 5.56\% O.A. at transformation ratios from 20\% to 80\%.

The performance deterioration of VeerCNN strongly suggests that CNN models are not robust to changes in the number of vertices in polygons. This can be attributed to several factors, including the data encoding method used in VeerCNN, which represents polygons as fixed-length 1D sequences with zero padding. Simplifying polygons by removing trivial vertices alters the length of these sequences, which not only dilutes the geometric information through zero padding but also disrupts the explicit encoding of vertex adjacency within the 1D sequences.

In comparison, we observe only minor performance drops in general set-based and graph-based learning models. This can be attributed to the flexibility of set and graph representations of polygons, which allow these models to accept input polygons of varying lengths. Specifically, graph-based message-passing networks (i.e., PolyMP and PolyMP-DSC) effectively learn local geometric features from neighboring nodes using a message-passing mechanism, followed by permutation-invariant pooling operations (e.g., max-pooling), as shown in Eq.~\ref{eq:mpn}. A similar pooling operation is utilized in set-based learning models \citep{zaheer2017deep,lee2019set}.

\subsection{Results on OSM Dataset}

\begin{table}[htbp]
\small
\centering
\caption{Fine-tuned test performance on the OSM dataset. OSM-O and OSM-R indicate the classification accuracy of models tested on original and rotated \& reflected samples, respectively. Overall accuracy is reported as OSM O.A.}
\begin{tabular}{lcccc}
\toprule
\textbf{Model} & \textbf{OSM-O Acc.} & \textbf{OSM-R Acc.} & \textbf{OSM O.A.} & \\
\midrule
\shortstack{VeerCNN} & 86.30 & 72.44 & 79.36 & \multirow{8}{*}{\rotatebox[origin=c]{0}{\shortstack{Normalised \\ coordinates  \\ $\in (-1, 1)$}}} \\
\shortstack{DeepSet} & 94.79 & \textbf{82.31} & \textbf{88.62} \\
\shortstack{SetTransformer} & 89.10 & 68.73 & 78.90 \\
\shortstack{GCAE}& 95.23 & 78.15 & 86.94 \\
\shortstack{NUFTSpec} & 88.02 & 60.66 & 74.26 \\
\shortstack{DSC-NMP} & 87.58 & 66.73 & 77.14 \\
\rowcolor{lightgray} \shortstack{PolyMP} & \textbf{96.59} & 81.03 & 88.58 \\
\rowcolor{lightgray} \shortstack{PolyMP-DSC} & 95.75 & 78.63 & 87.20 \\
\midrule
\shortstack{VeerCNN} & 42.23 & 42.09 & 42.16 & \multirow{8}{*}{\rotatebox[origin=c]{0}{\shortstack{Original \\ coordinates}}}\\
\shortstack{DeepSet} & 22.00 & 21.65 & 21.84 \\
\shortstack{SetTransformer} & 35.62 & 35.30 & 35.46\\
\shortstack{GCAE} & 22.12 & 22.32 & 22.22 \\
\shortstack{NUFTSpec} & \textbf{88.02} & \textbf{61.38} & \textbf{74.42} \\
\shortstack{DSC-NMP} & 31.81 & 33.23 & 32.52 \\
\rowcolor{lightgray} \shortstack{PolyMP} & 50.64 & 39.86 & 46.20 \\
\rowcolor{lightgray} \shortstack{PolyMP-DSC} & 44.19 & 36.78 & 40.68 \\
\bottomrule
\end{tabular}
\label{tab:osm}
\end{table}

The results in Table~\ref{tab:osm} show the test accuracy of fine-tuned classifiers on real-world building footprints from the OSM dataset, using feature encoders pre-trained on the Glyph dataset (80\% subset). These fine-tuned models were evaluated on building footprints with both normalized and original coordinates.

PolyMP achieves an accuracy of 88.58\% for building footprints with normalized coordinates and 46.20\% for footprints with original coordinates. PolyMP-DSC demonstrates slightly lower test accuracy, with 87.20\% for normalized coordinates and 40.68\% for original coordinates, but still outperforms other baseline methods, except for DeepSet. This can be attributed to the ability of both models to compute local geometric features via message-passing networks, reducing the impact of absolute positioning while preserving polygon geometry. DeepSet performs best, with 88.62\% O.A. on normalized coordinates and 82.31\% O.A. on rotated and reflected buildings, making it the top performer in some cases.

The NUFTSpec model, which transforms polygon geometries into consistent spectral feature vectors via Non-Uniform Fourier Transforms, demonstrates stable performance at approximately 74\% O.A. across both normalized and original coordinates. This stability stems from a series of affine transformations (e.g., scaling and translation) applied to input polygons during feature transformation, ensuring that each polygon is projected into a unit space and positioned within the same relative coordinate system, as discussed in \citep{mai2023towards}.

\begin{figure}[htbp]
\centering
    \subfloat[OSM-O]{%
    \resizebox*{1\textwidth}{!}{\includegraphics{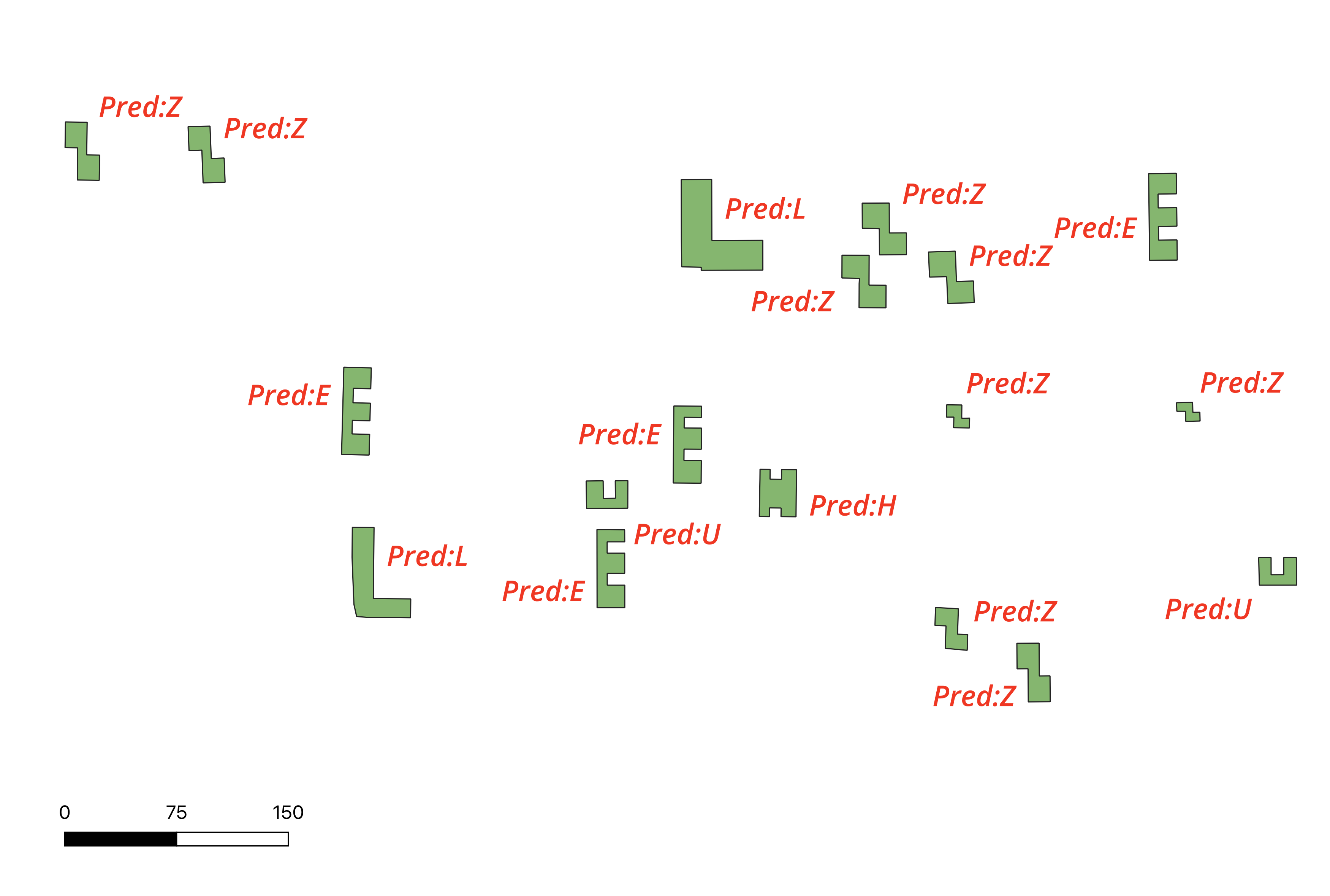}}}
\hfill
    \subfloat[OSM-R]{%
    \resizebox*{1\textwidth}{!}{\includegraphics{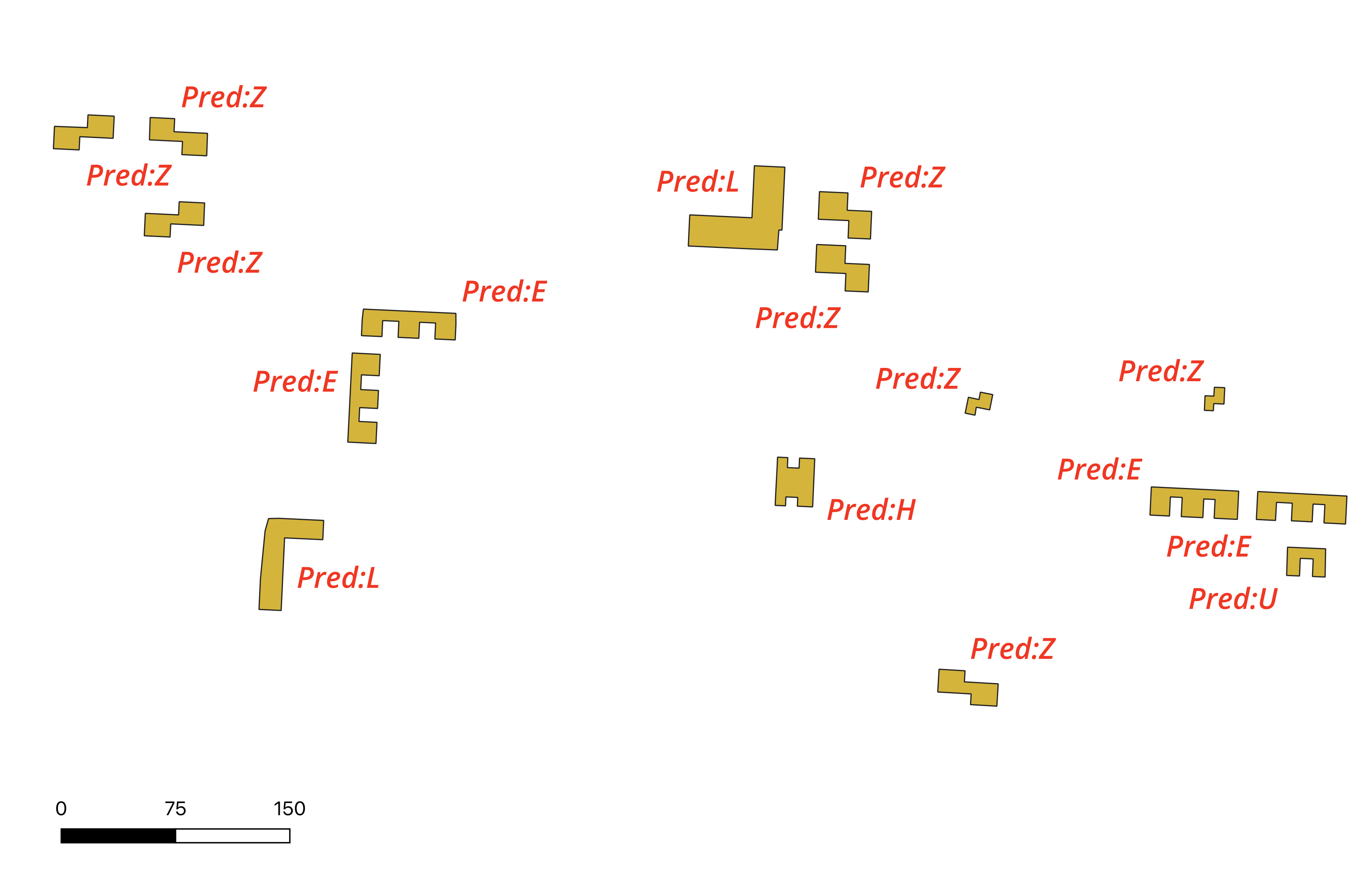}}}
\hfill
\caption{Shape classification of building footprints using PolyMP on a selected region of the OSM dataset \citep{yan2021graph}. Model predictions are highlighted in red.}
\label{fig:OSM_map}
\end{figure}

In Fig.~\ref{fig:OSM_map}, classification results of PolyMP on building footprints from the two OSM subsets (OSM-O and OSM-R) are visualized. The results show PolyMP's robustness in classifying building footprints of varying sizes and geometric shapes, leveraging the geometric invariant features learned through message-passing networks.

These findings underline the effectiveness of graph message-passing models, particularly PolyMP and PolyMP-DSC, in handling both normalized and original coordinate systems and achieving strong performance across various data transformations.

\section{Discussion}
\label{disucssion}

\begin{figure*}
\centering
    \subfloat[Sheared M glyph.\label{fig:feat_m1}]{%
    \resizebox*{1\textwidth}{!}{\includegraphics{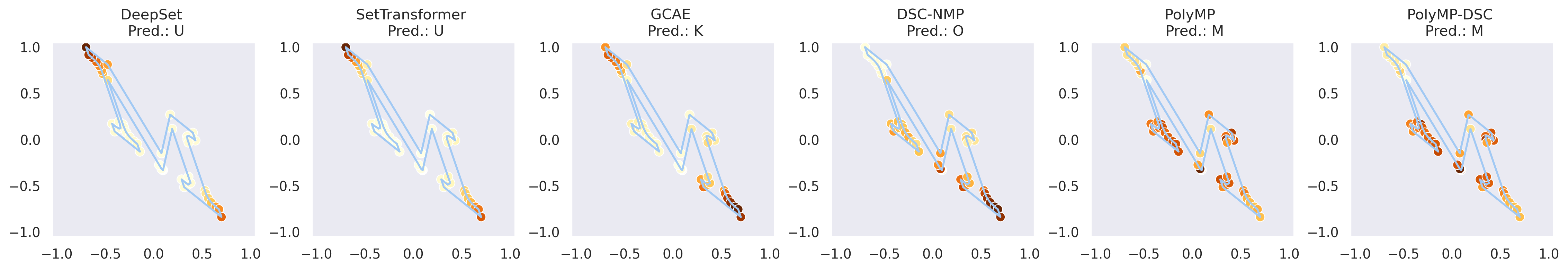}}}
\hfill
    \subfloat[Rotated M glyph.\label{fig:feat_m2}]{%
    \resizebox*{1\textwidth}{!}{\includegraphics{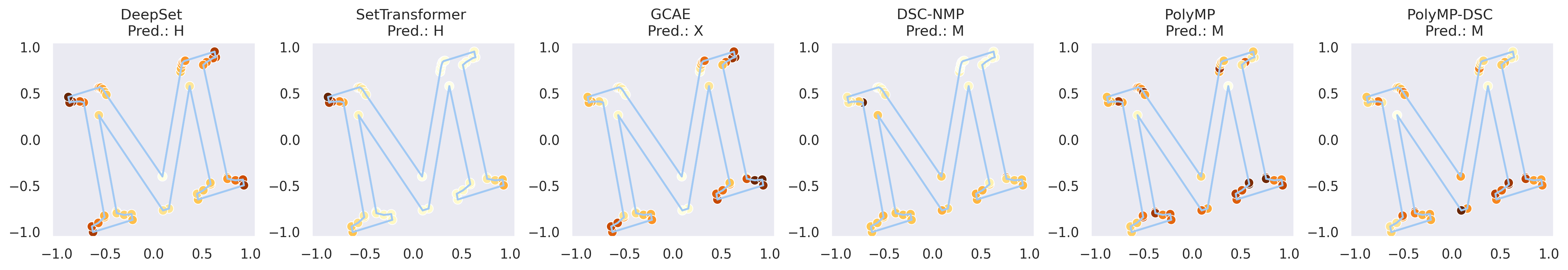}}}
\hfill
    \subfloat[Scaled W glyph.\label{fig:feat_w2}]{%
    \resizebox*{1\textwidth}{!}{\includegraphics{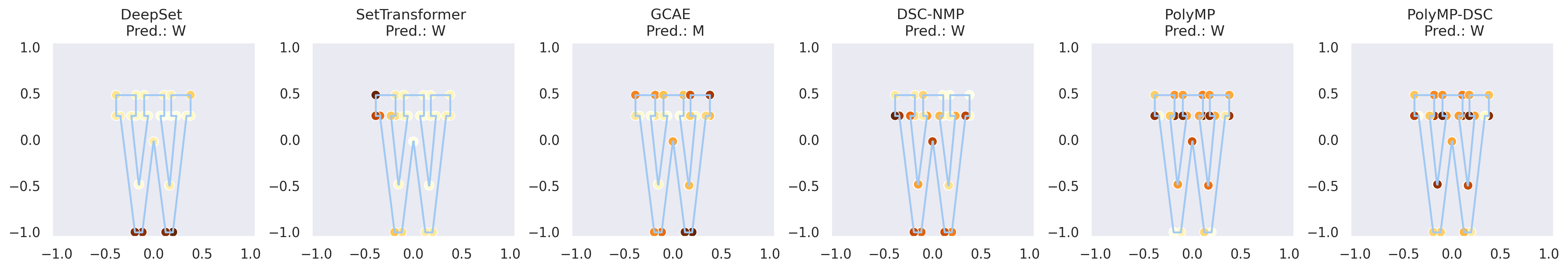}}}
\hfill
    \subfloat[Rotated W glyph.\label{fig:feat_w1}]{%
    \resizebox*{1\textwidth}{!}{\includegraphics{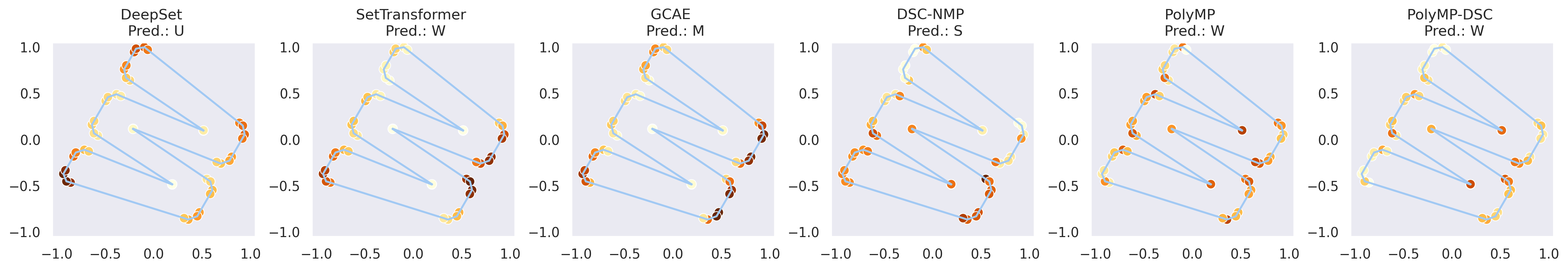}}}
\hfill
    \subfloat[Sheared R glyph.\label{fig:feat_r1}]{%
    \resizebox*{1\textwidth}{!}{\includegraphics{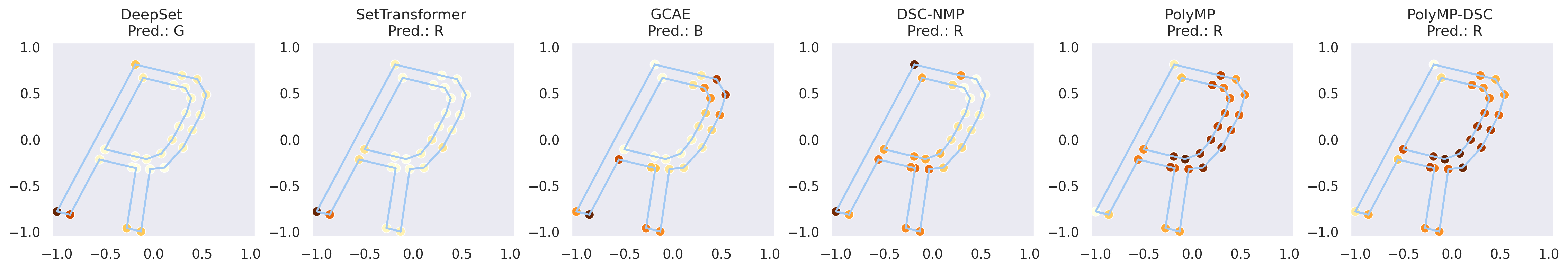}}}
\hfill
    \subfloat[Original Q glyph.\label{fig:feat_q1}]{%
    \resizebox*{1\textwidth}{!}{\includegraphics{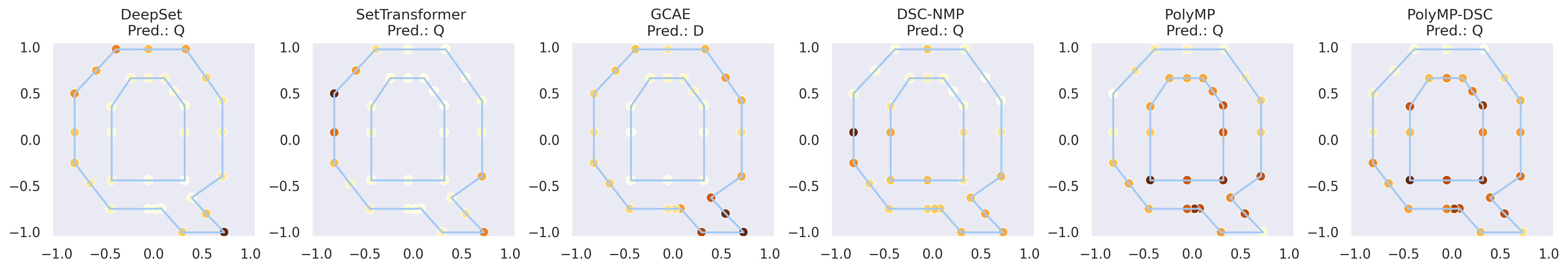}}}
\caption{Feature maps to visualize geometric features learned by individual models to classify input geometries to output categories. The darker the color, the more salient are the features learned. Feature values are normalised $\in [0, 1]$.}
\label{fig:feat_vis}
\end{figure*}

\begin{figure}
\centering
    \subfloat[Glyph dataset.\label{fig:failure_glyph}]{%
    \resizebox*{1\textwidth}{!}{\includegraphics{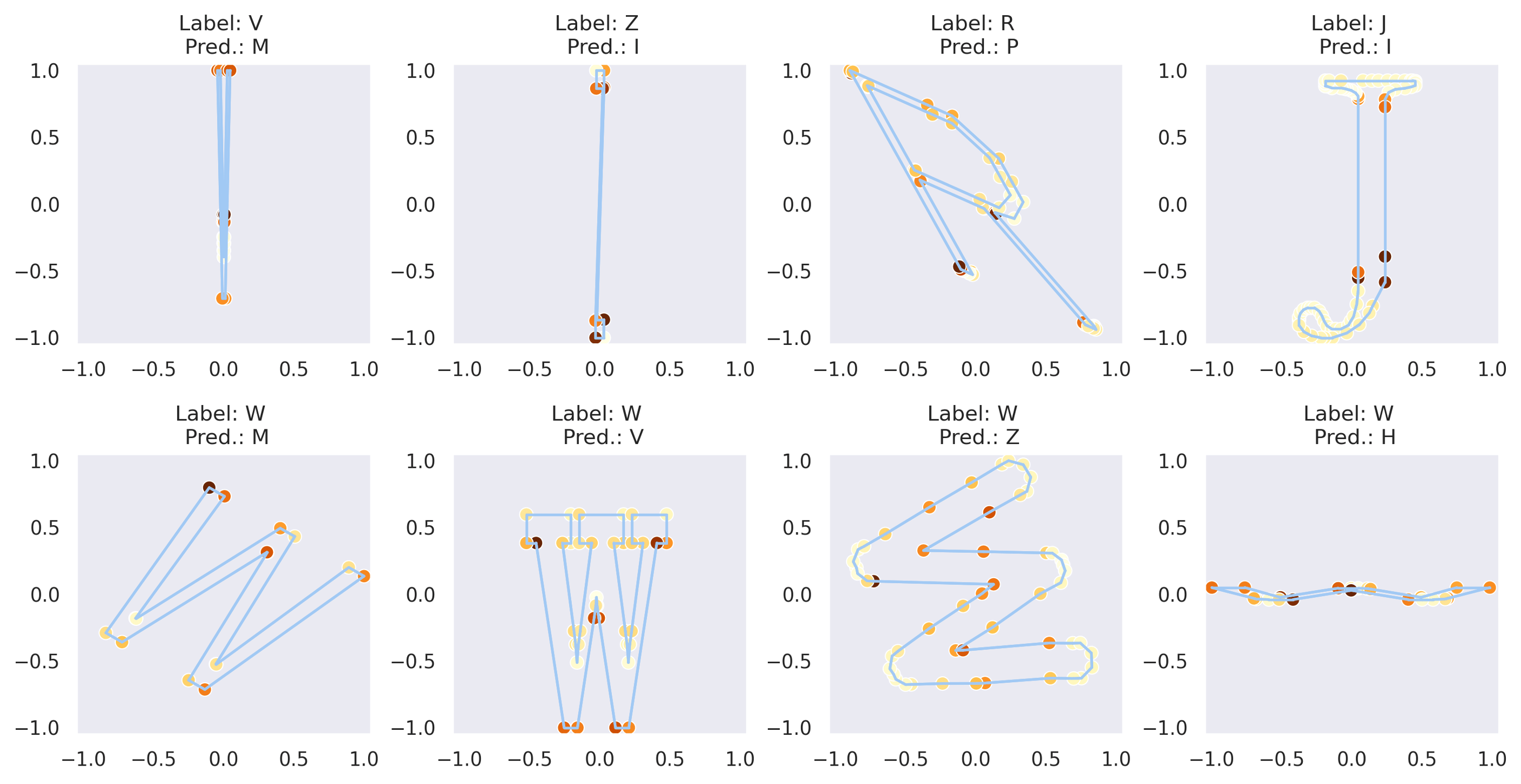}}}
\hfill
    \subfloat[OSM dataset.\label{fig:failure_osm}]{%
    \resizebox*{1\textwidth}{!}{\includegraphics{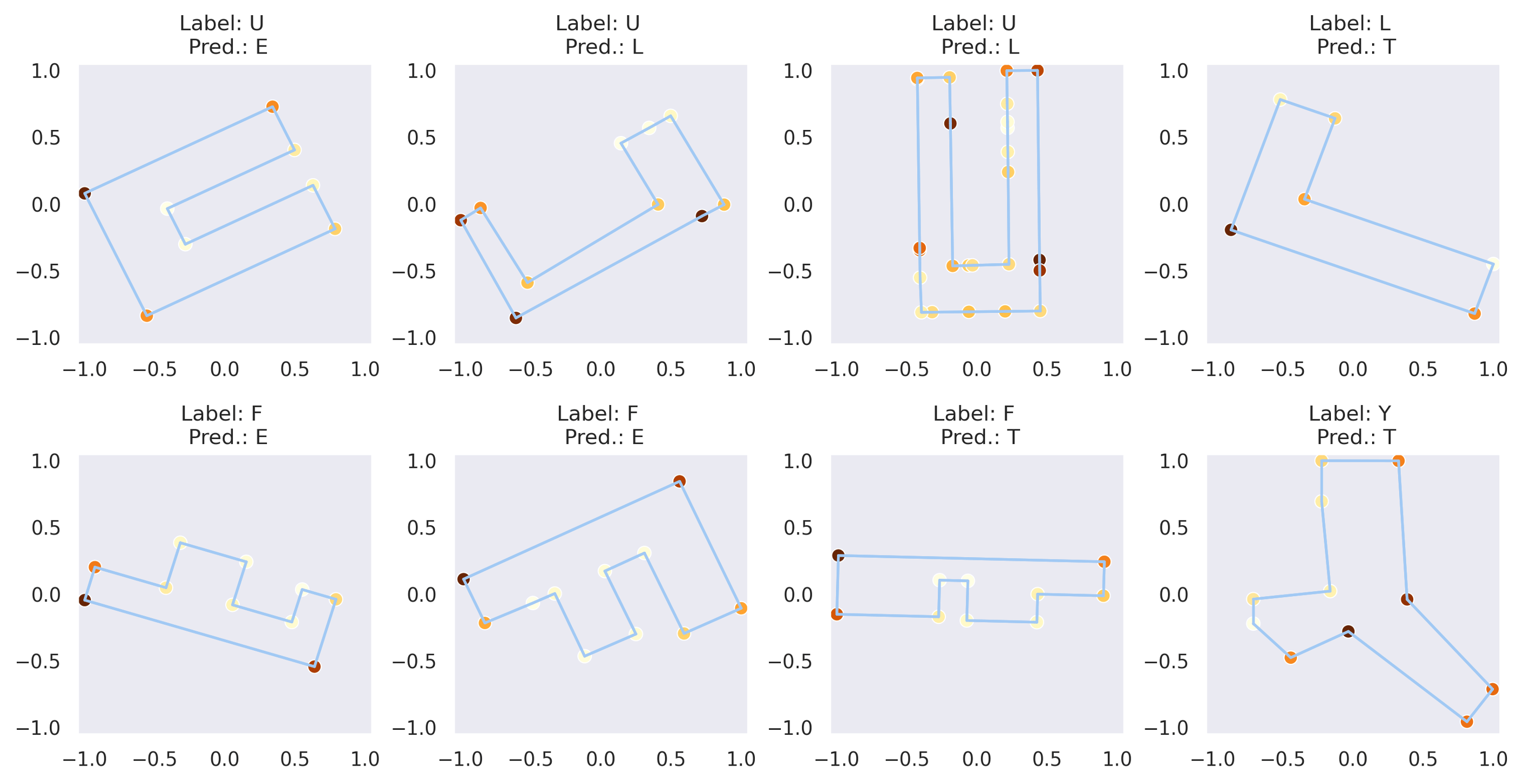}}}
\caption{Failure cases of PolyMP on Glyph and OSM datasets.}
\label{fig:failure}
\end{figure}

\subsection{Model Expressivity}
To interpret the expressivity of different polygon encoding and learning models, we visualize the feature maps of models tested on original and geometrically transformed samples from the Glyph dataset in Figure~\ref{fig:feat_vis}.

Figures~\ref{fig:feat_m1} and \ref{fig:feat_m2} illustrate the salient feature maps and class predictions of set-based (DeepSet, Set Transformer) and graph-based (GCAE, DSC-NMP, PolyMP, and PolyMP-DSC) learning models on geometrically transformed samples. Notably, PolyMP and PolyMP-DSC correctly predict the class label of the sheared and rotated glyph $M$, whereas DeepSet, Set Transformer, GCAE, and DSC-NMP make incorrect predictions. By comparing the feature maps in Figures~\ref{fig:feat_w2} and \ref{fig:feat_w1}, we observe that PolyMP and PolyMP-DSC effectively capture the global structure or \enquote{skeleton} of polygons by identifying salient, non-trivial vertices along the boundary. The message-passing encoder, coupled with a local pooling function (i.e., max-pooling), extracts geometric features from local neighborhoods, where non-trivial vertices (darker color) are distinguished within clusters of trivial vertices (lighter color). This enables the models to generate robust embeddings of polygons, even under geometric transformations. In contrast, set-based models and GCAE learn sparser geometric representations, capturing regional features in node clusters while neglecting relational information between more distant nodes. This limitation arises from their reliance on set representations, which inherently discard connectivity information between vertices.

Figures~\ref{fig:feat_r1} and \ref{fig:feat_q1} illustrate the feature maps for polygons with holes. DeepSet and Set Transformer primarily capture geometric features from non-trivial vertices along the exterior boundary, while largely ignoring those within the interior. Although set-based models can encode polygons of varying lengths and produce permutation-invariant feature representations, their lack of connectivity modeling restricts their ability to learn geometric structures in polygons with holes. Similarly, the graph convolution model GCAE struggles to flexibly capture local geometric features, as discussed in previous sections. In contrast, PolyMP and PolyMP-DSC effectively capture geometric features from both the exterior and interior boundaries of polygons. This is achieved through graph message-passing encoders, where geometric features (i.e., the \enquote{message}) are exchanged only between directly connected nodes within the same connected component of a graph—defined in this case by the polygon’s linear rings. As a result, PolyMP-based models can better preserve structural relationships in polygons with complex topologies.

\subsection{Limitations}
Despite its strong performance, our proposed method still encounters misclassification limitations. Figure~\ref{fig:failure} highlights instances where PolyMP fails to classify samples correctly in the Glyph and OSM datasets.

On synthetic data, misclassifications predominantly occur in cases where excessive geometric transformations—such as heavy shearing and anisotropic scaling—severely distort the shape’s structure, rendering it unrecognizable: rotated W shapes (Fig.\ref{fig:failure_glyph}) are sometimes mistaken for M or even Z, while heavily sheared W shapes may be misclassified as H or V.

Similarly, in the OSM dataset, PolyMP and PolyMP-DSC occasionally misclassify building polygons with similar geometric structures, such as E, L, and U shapes (Fig.~\ref{fig:failure_osm}). This can be attributed to the model’s reliance on local geometric features, which may not sufficiently distinguish between structures with high shape similarity.

Considering the characteristics of the two datasets (Sec.~\ref{glyph} and \ref{osm}), PolyMP and PolyMP-DSC trained on the Glyph dataset primarily learns locally salient geometric features through message-passing encoders. These features are particularly effective for serif glyphs, which contain distinctive decorative stroke endings. However, when excessive geometric transformations distort these key features, the model struggles to preserve label consistency, leading to misclassification among shapes with similar structures.

\section{Conclusion}
\label{conclusion}

In this study, we investigated the challenge of geometric-invariant shape classification from 2D vector polygons, a common but underexplored data format in spatial analysis. Through comprehensive empirical evaluation, we demonstrated that combining discrete, non-grid-like polygon representations with graph-based message-passing neural networks—PolyMP and its densely self-connected variant, PolyMP-DSC—enables the learning of robust and expressive geometric features.

Our findings show that graph representations of polygons, which encode both geometric structure and vertex connectivity, significantly outperform sequence- and set-based encodings in terms of transformation invariance and generalization. In particular, PolyMP-DSC enhances feature propagation by incorporating dense self-connections, resulting in improved classification performance and robustness to structural perturbations such as trivial vertices. These properties are crucial for geospatial tasks where object shapes may undergo rotation, scaling, shearing, or digitization noise.

Importantly, our approach aligns well with practical geospatial workflows: polygon-to-graph conversion is natively supported by many GIS libraries, enabling seamless adoption of our models in downstream applications such as automated cartographic generalization, building footprint recognition, and road geometry analysis.

Nevertheless, current limitations remain. Both PolyMP and PolyMP-DSC occasionally struggle with locally distinctive features that vary across shape classes or domains. To address this, future work will explore the integration of complementary geometric-invariant descriptors—such as spectral features (e.g., NUFT) and local shape attributes (e.g., turning angles, curvature, or radii of arc segments)—to further improve model robustness and invariance.

Overall, this work highlights the importance of representation choice in spatial deep learning. By designing models with inductive biases that reflect geometric structure and transformation invariance, we move closer to human-like perception of shape in geospatial domains. Beyond classification, our findings open promising directions for extending graph-based learning to polygon regression, spatial clustering, and topological inference in large-scale vector datasets.
\backmatter

\section*{Declarations}

\subsection*{Data availability.} The data and code supporting the findings of this study are available at \url{https://github.com/zexhuang/PolyMP}.

\subsection*{Conflicts of interest.} This paper has been approved by all co-authors. The authors have no competing interests to declare that are relevant to the content of this article.

\subsection*{Funding.} No funding was obtained for this study.

% \begin{appendices}

% \section{Section title of first appendix}\label{secA1}

% An appendix contains supplementary information that is not an essential part of the text itself but which may be helpful in providing a more comprehensive understanding of the research problem or it is information that is too cumbersome to be included in the body of the paper.




% \end{appendices}
\clearpage
\bibliography{sn-bibliography}% common bib file

\end{document}